\documentclass[11pt,a4paper]{article}
\usepackage[hyperref]{naaclhlt2019}
\usepackage{times}
\usepackage{latexsym}

\usepackage{url}

\usepackage{amsfonts}
\usepackage{amsmath}
\usepackage{graphicx}

\usepackage{color}
\newcommand{\das}[1]{\textcolor{red}{//\emph{das:} #1//}}

\newcommand{\sketch}{\textcolor{red}{//Rough draft only.//}}

\usepackage{enumitem}

\aclfinalcopy 

\newtheorem{definition}{Definition}

\title{Language Tasks and Language Games: On Methodology\\ in Current Natural Language Processing Research}

\author{David Schlangen\\
  Computational Linguistics / Department of Linguistics\\
  University of Potsdam, Germany\\
  {\tt david.schlangen@uni-potsdam.de} \\}

\date{}

\begin{document}
\maketitle
\begin{abstract}
  ``This paper introduces a new task and a new dataset'', ``we improve the state of the art in X by Y'' -- it is rare to find a current natural language processing paper (or AI paper more generally) that does \emph{not} contain such statements. What is mostly left implicit, however, is the assumption that this necessarily constitutes progress, and what it constitutes progress towards. Here, we make more precise the normally impressionistically used notions of \emph{language task} and \emph{language game} and ask how a research programme built on these might make progress towards the goal of modelling general language competence.
\end{abstract}

\section{Introduction}
\label{sec:intro}


Recently, seemingly ever other natural language processing paper introduces a new task and a new dataset.\footnote{%
  Not quite, but not very far. Looking at the 2018 long and short paper proceedings of ACL and EMNLP, we get 94 hits for ``introduce new dataset'', 20 hits for ``introduce new corpus'', and 101 hits for ``introduce new task''.
}
We join some other recent papers 
\cite[e.g.][]{Yogatama2019} in asking whether there is any coherence to this research approach, under which conditions it can lead to progress, and towards what. What we do differently, however, is to look at the fundamental assumptions behind this approach. We try to define central notions, in order to be able to discuss the structure of the typically only implicit formulated approach more clearly.




In our argumentation, we distinguish between \emph{language tasks}, such as for example ``describe this image'', or ``translate this sentence''---that is, single-step tasks that involve in an essential way natural language material, but not necessarily \emph{only} language material---; \emph{micro worlds}, which are environments that produce disinterested responses to actions, thereby possibly simulating the behaviour of independently existing systems; and \emph{dialogue games} as repeated and connected language tasks, which these environments enable. We define these notions first and think about general ways of evaluating their relevance.
We close with some tentative recommendations for how to connect individual modelling contributions with the larger enterprise of modelling language processing.

\section{Tasks, Worlds, and Games}
\label{sec:defs}

\begin{figure*}[ht]
  \centering
  \includegraphics[width=.8\linewidth]{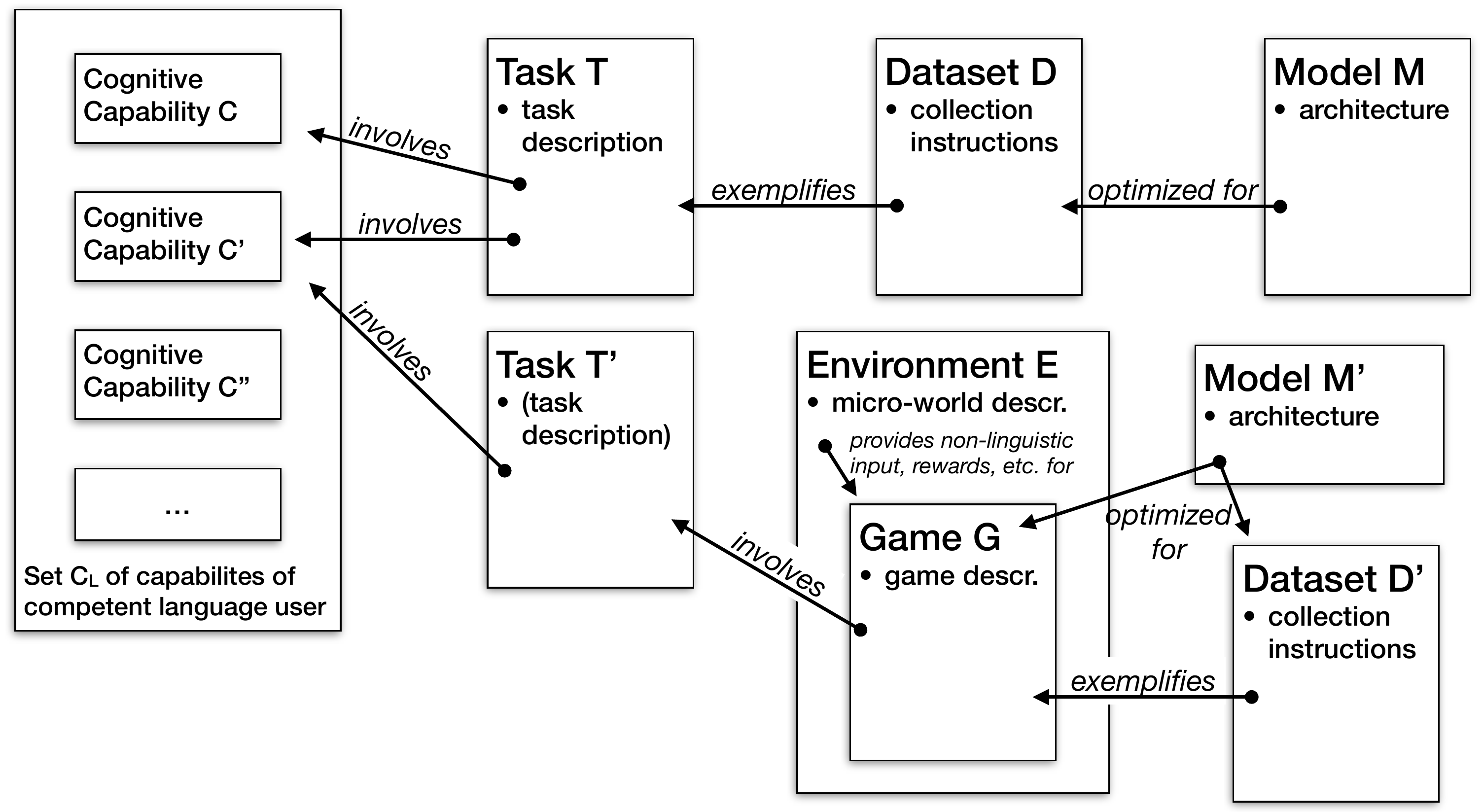}
  \vspace*{-1ex}
  \caption{The structure of relations between the research objects \emph{model}, \emph{dataset}, \emph{task}, \emph{game}, \emph{environment}, \emph{cognitive capability}.}
  \label{fig:tag}
  \vspace*{-2ex}
\end{figure*}



\subsection{Tasks}
\label{sec:tasks}

A \emph{language task} is a mapping between an \emph{input space} and an \emph{output} or \emph{action space}, at least one of which contains natural language expressions.\footnote{%
  A formal definition of this and the other notions is given in the Appendix.
}
The mapping has to conform to a \emph{task description}, which
is typically given only informally, making reference to theoretical or pre-theoretical constructs external to the definition, such as ``translation'' or ``is true of''. We call this an \emph{intensional description}. Often, a task is also specified \emph{extensionally} through the provision of a \emph{dataset} of examples of the mapping (that is, pairs of state and action), $\mathcal{X} = \{(x_1,y_1), \dots, (x_m, y_m)\}$, where the assumption is that $(x,y) \in \mathcal{X} \; \rightarrow \; y = \mathcal{L}(x)$ ($\mathcal{L}$ being the task mapping).


This very general definition essentially covers much, if not all of natural language processing. For example, translation can be seen as a language task where the state space consists of expressions in one language, the action space of expressions in another language, and the task description
is that in each pair in the mapping, the second element be a translation of the first. We can further distinguish \emph{understanding tasks}, where the mapping requires demonstration of sensitivity to language meaning (however that is to be further defined); \emph{interpretation tasks}, where the input space contains language expressions that are to be ``understood''; \emph{generation tasks}, where the output does (with a given task potentially being both an interpretation and a generation task); \emph{reference tasks}, where the understanding is shown by relating linguistic and non-linguistic material, and \emph{inference tasks}, where linguistic material is related.


\subsection{Worlds}
\label{sec:worlds}

The language competence of humans plays out in \emph{repeated} task,
not single-step ones as described in the previous section, and it plays out 
in contexts where language use is embedded in a non-linguistic context. To study such repeated, situated games, much recent work has made use of environment simulators that compute reactions to actions performed within them, in accordance to the assumed (or actual) rules of the domain they represent.\footnote{%
  See for example \cite{Savva2019,Adams2012,Johnson2016:malmo,Urbanek2019,Baroni2017:commai,Xia2018,Yan2018,Misra2018,Cote2018,Bennett2018,Anderson2018,Savva2017,Gordon2017,Brodeur2017,Chang2017a,Janarthanam2011,Baroni2017,Byron2007,Yamauchi2013}.
  }
Such environments can again be described as mappings, in this case from an action to an environmental response (a state), where again the mapping conforms to a description of which real-world counterpart it is intended to model, and how.\footnote{%
  In this desire to model the relevant aspects of a domain, and in the assumption that from dealing with a simulated environment transferable knowledge about dealing with the original enviroment can be achieved, this approach is reminiscent of the AI microworlds of the 1970s---``we see solving a problem often as getting to know one's way around a `micro-world' in which the problem exists.'' \cite{MinskyPapert1972}---and perhaps 
  susceptible to similar kinds of critiques as these attempts \cite{Dreyfus1981,marr:vision}.
}


\subsection{Games}
\label{sec:games}

An \emph{interaction game} is a setting where \emph{players} can produce \emph{actions} (a special kind of which can be \emph{messages}), possibly regulated by some regime on when they can do this and who can observe them. A priviledged, but disinterested player \emph{Nature} can respond to those actions, by providing game-relevant information (and interfacing with the environment in which the game is embedded). Again, we assume that there is an informal \emph{game description} which specifies which, if any, otherwise existing activity the game is meant to model. A \emph{language game} is an interaction game that embeds language tasks which govern the actions of the players. For example, one player asking and the other player answering questions would be a language game that poses repeated language tasks to the players.




\ \\
Summarising this section, Figure~\ref{fig:tag} shows the structure of relations between the notions introduced here: Models relate to Tasks, via Datasets, Games are realised in Environments. \emph{Capabilities} will be discussed below.

\section{What makes a good task, world, game?}
\label{sec:eval}


Let's now assume we encounter a paper that proposes a new dataset, language task, microworld, or language game. How can we evaluate the contribution that is made?

\subsection{Tasks}
\label{sec:taskeval}



\paragraph{... and datasets}

As mentioned, tasks are often exemplified by the provision of a dataset of examples of the task being executed by agents that are assumed to be capable of doing so---typically, human participants in experiments or data collection efforts. Evaluating such datasets in itself is relatively straightforward. First, it should be \emph{verified}, which is to check whether the provided input/output pairs can indeed be judged correct relative to the task (in its intensional description). If the examples are collected specifically for the purpose of exemplifying the task, this is the process of controlling annotation, and standard methodologies exists \cite{ArtPoesio:kappa}.

\emph{Validating} a dataset is a less formalised process. It comprises arguing that the dataset indeed exemplifies the task intension well. For example, pairs only of images of giraffes and sentences describing them would arguably not exemplify the general task of \emph{image description} very well (even if the descriptions are accurate), while perhaps exemplifying the task of \emph{giraffe image description}.

Another way to evaluate datasets is by providing a model of the task learned on parts of it, and testing it on the remaining part (for which a comparison, or \emph{loss}, function on input/output pairs must be provided as well). If a model can ``solve'' the dataset even when not given information that for theoretical or pre-theoretical reasons is seen to be crucial, the dataset can be considered an unsatisfactory exemplification of the task. E.g., in a \emph{visual (polar) question answering} setting \cite{VQA2015}, if in a dataset all and only the expressions that mention giraffes are true, a model would not need to take the images into account at all to perform well (as it would just need to detect the presence of giraffe-related words), which would be evidence that the dataset is deficient relative to the task description.

\paragraph{... in themselves}

How can a task in itself be motivated and evaluated? This is easy, if it has a direct value to a consumer (such as translation presumably has), which can be measured. If the consumer is a computer system that processes the output of the task further, the burden of evaluation is simply shifted to the system as a whole. If the interest is in replicating with a theoretically motivated model performance characteristics of humans attempting the task, the task can be evaluated for its power helping distinguish between different modelling choices.

A recent trend, however, has been to motivate tasks in a different way, neither via their inherent practical use, nor as answering questions about language processing as implemented in humans. The argument roughly goes as follows (even if typically only made implicitly): To be good at task $T$, an agent must possess a set $C_T$ of capabilities (of representational or computational nature). If the $c \in C_T$ are capabilities that competent language users 
can be shown or argued to possess and make use of in using language---let's call the set of these capabilities of a competent language user $C_L$, so that $C_T \subseteq C_L$--- then being able to model these capabilities (via modelling the task) results in progress towards the ultimate goal, which is to model competent language use. And hence, any task $T$ that comes with an interesting set $C_T$ is a good task.\footnote{%
  To give some examples of informal versions of this argument, and chosing papers more or less randomly, here are some quotes (typically from the introduction sections of their respective papers):

  From the paper that introduced the \emph{visual question answering} task \cite{VQA2015}: ``What makes for a compelling “AI-complete” task? We believe that in order to spawn the next generation of AI algorithms, an ideal task should (i) require multi-modal knowledge beyond a single sub-domain (such as CV) and (ii) have a well-defined quantitative evaluation metric to track progress. [\dots] Open-ended questions require a potentially vast set of AI capabilities to answer -- fine-grained recognition (e.g., “What kind of cheese is on the pizza?”), object detection (e.g., “How many bikes are there?”), activity recognition (e.g., “Is this man crying?”), knowledge based reasoning (e.g., “Is this a vegetarian pizza?”), and commonsense reasoning (e.g., “Does this person have 20/20 vision?”, “Is this person expecting company?”).''

  About the \emph{natural language inference} problem, and attempting at least an implicit structuring of the space of capabilities, \citet{Condoravdi2003} write: ``The ability to recognize such semantic relations is clearly not a \emph{sufficient} criterion for language understanding: there is more to language understanding than just being able to tell that one sentence follows from another. But we would argue that it is a minimal, \emph{necessary} criterion.'' 

  \citet{Williams2018}, on the modern version of this task: ``The task of natural language inference (NLI)
is well positioned to serve as a benchmark task for research on NLU. [\dots] In particular, a model must handle phenomena like lexical entailment, quantification, coreference, tense, belief, modality, and lexical and syntactic ambiguity.''

}

Under what conditions does this argument work? First of all, the assumed connection to the set of capability must indeed be there. We have already seen a way to \emph{challenge} a claimed connection, namely through providing a model that can ``solve'' a given task (via a dataset) while not having access to information that should be involved in the capability. (Although this challenge in the first instance only targets the dataset and not the task itself.)
Secondly, following usual scientific methodology \cite{popper:logik}, we can rank the worth of an instantiation of this argument by how precisely the capability is specified, from the trivially correct ``task T involves the capability to do task T'' to a statement that could be wrong (and hence involves other theoretical constructs), e.g.\ ``task T involves the capability to compute the syntactic structure of a natural language sentence''.

Furthermore, we can rank the motivation given for a task by how explicit it is in delineating the set of capabilities it involves, along two dimensions. In the one dimension (\emph{separability}), in the most extreme form, the claim would be that the set of capabilities $C_T$ is fully separated from $C_L \setminus C_T$, and hence there is no danger of overfitting solutions to $C_T$ in such a way as would be detrimental for the remaining capabilities.\footnote{%
  That is, unless the claim is that $C_T = C_L$; the use of ``AI-complete'' in the quote above from \citet{VQA2015} suggest that is something that not everyone shies away from.
} In the other dimension (\emph{exhaustivity}), the strongest claim would be that $T$ brings out all there is to $c \in C_T$, and that another task $T'$, insofar as it requires $c$ as well, could be handled by a model of $c$ built with only $T$ in mind. In the other extreme, we only have ``$c$ as required by $T$'', which does less to indicate progress beyond $T$.

As this discussion suggests, it seems difficult to properly motivate a task without relying, at least implicitly, on assumptions about how $C_L$ decomposes.

\subsection{Worlds}
\label{sec:worldeval}



We have introduced ``worlds'' (or environments, or simulators) above as the settings that enable repeated tasks (\emph{games}), and in parts their evaluation is connected to those. However, as the providers of \emph{inputs} to a task in reaction to its \emph{outputs}, where this mapping must also confirm to a description, we can also evaluate them qua environment.

First, the notions of \emph{validation} and \emph{verification} apply as well: An environment should match, as well as possible, its stated description of its relation to a real-world counterpart, and should be verified to be correct according to its specification. Environments like those needed for games like Chess and Go \cite[e.g.][]{Silver2017}, or Settlers of Catan \cite[e.g.][]{afentenos:etal:2012}, can fully model their intended real-world counterpart, whereas others can only be modelled aproximately
(e.g., ``the interior of a house, through which an agent moves, from the perspective of that agent'' --- as the body of the agent and the agent's awareness of it, arguably, is a part of the environment, this would entail modelling this as well).



A detailed list of desiderata for such ``artificial general intelligence'' environments is given by \citet{Adams2012} (see also \citet{Baroni2017}): ``C1. The environment is complex, with diverse, interacting and richly structured objects. C2. The environment is dynamic and open. C3. Task-relevant regularities exist at multiple time scales. C4. Other agents impact performance. C5. Tasks can be complex, diverse and novel. C6. Interactions between agent, environment and tasks are complex and limited.
C7. Computational resources of the agent are limited. C8. Agent existence is long-term and continual.'' While this list mixes what we separate as demands on environments and on games set in them, it should be useful to evaluate proposed environments.

In analogy to what we observed for tasks, it seems that 
trying to make progress through modelling tasks in simulated worlds entails making another separability hypothesis, which assumes that the natural 
competence of handling the world as a whole is separable into handling various parts of it, which can be ``knit'' together to form the whole.\footnote{%
  ``[W]e feel [the micro-worlds] are so important that we plan to assign a large portion of our effort to developing a collection of these micro-worlds and finding how to embed their suggestive and predictive powers in larger systems without being misled by their incompatibility with literal truth.
  [\dots] In order to study such problems, we would like to have collections of knowledge for several `micro-worlds', ultimately to learn how to knit them together.'' \cite{MinskyPapert1972}.
}





\subsection{Games}
\label{sec:gameeval}



What makes for an interesting language game? First we note that games seem to be less well exemplified by datasets than (non-game) tasks are, as for them the relation between an input and an output is much less constrained, and the output can be a sequence of actions rather than a simple one. To give an example, in the language-navigation task \cite[e.g.][]{Anderson2018,Ma2019}, while the input is a single datum (a verbal description of a goal location), the output is a sequence of navigation actions.\footnote{%
  On the role of environments/games vis-\`a-vis datasets, \citet{Savva2019} state their belief that ``simulators will assume the role played previously by datasets''; no further argumentation is given to support this belief, however.
}

As the site for repeated and connected language tasks, we can evaluate a game again for how it connects to language capabilities, and for how the game setting improves over a (non-repeated) task setting. For example, some previous work has shown that language production under (interactive) task constraints is different from null-context language production \cite{ilinykh:inlg18,DaSilvaRocha2016}.

Finally, it seems that for language games there is a natural supremum, which would be ``unrestricted situated language interaction''. We can then also evaluate proposed tasks for how close they come to this ultimate form, for some definition of ``closeness''.

\begin{figure*}[ht]
  \centering
  \includegraphics[width=.9\linewidth]{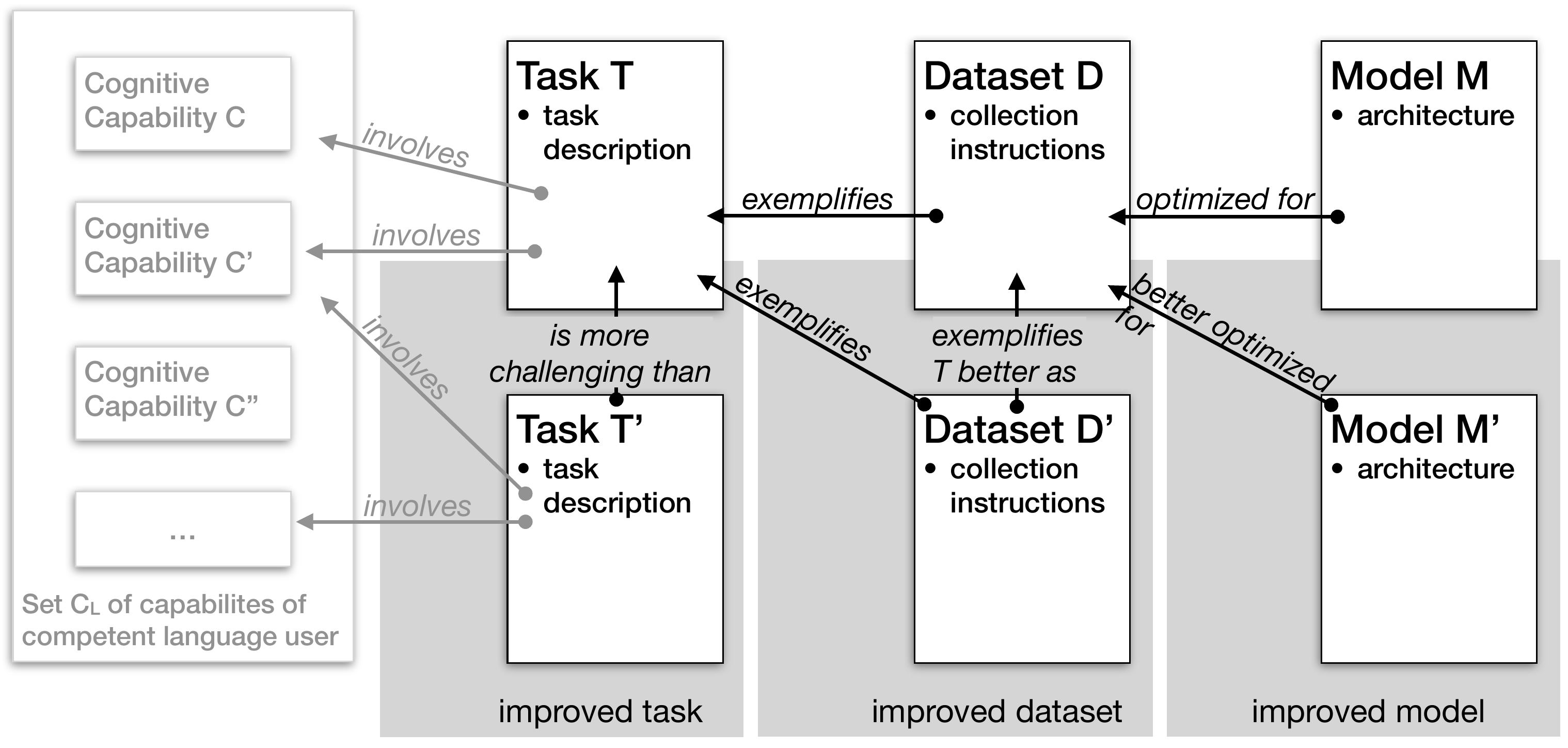}
  \caption{Three ways of making progress: improving models, improving datasets, improving tasks. (Not shown: devising models that handle more than one dataset.)}
  \label{fig:prog}
\end{figure*}

\section{Making Progress}
\label{sec:prog}

\begin{quote}
  \emph{{\"U}berhaupt hat der Fortschritt das an sich, da{\ss} er viel gr\"o{\ss}er ausschaut als er wirklich ist.} (It is in the nature of progress that it appears much greater than it actually is.) \textsc{Nestroy}, via \citet{Witt:PU}
\end{quote}


\noindent
Research is an incremental enterprise, and new tasks, datasets, models, environments, and games are introduced in a context of existing ones. We can now distinguish several modes of making progress in the general project, by relating new proposals to existing ones. 

\subsection{Better Models}
\label{sec:mod}

This is the mode of progress for most of the current work in NLP / AI, and it is also the one that needs the least argumentative support: If a model trained on a given dataset performs better, in terms of the same pre-defined metrics, than a previous model, then progress has been made. Or would that it were so simple: Even in this setting, considerations of computing power spent should arguably also factor, and how to account for use of larger datasets (for example in pre-training), when judging models that achieve better results.\footnote{%
  See e.g.\ the recent discussion on \url{https://hackingsemantics.xyz/2019/leaderboards/}.
}
And ultimately, the worth of progress in modelling a particular task rests only on the worth of the task itself.

A different way in which models can improve over others is in how well they are suited for transfer to other tasks (see \citet{Ruder2019} for a recent overview of such transfer-learning). 

\subsection{Better Datasets}
\label{sec:data}

We have discussed the relation between tasks and datasets above. If a dataset can be shown to be successfully modelled even in the absence of information that is deemed criticially involved in the capability of interest, it can be considered invalid relative to the task description, and, if the interest in the task is kept, a new dataset must be found.

The task of visual question answering provides an interesting example case of such a development. After \citet{VQA2015} introduced the first large scale dataset for this task, it quickly became clear that this dataset could be handled competitively by models that were deprived of visual input \cite[``language bias'', as noted e.g.\ by][]{Jabri2016}. This problem was then addressed by \citet{Goyal2017} with the construction of a less biased (and hence more valid) corpus. Targetting the set of capabilities involved in the task, \citet{Andreas2016a} noted that ``questions in most existing natural image datasets are quite simple, for the most part requiring that only one or two pieces of information be extracted from an image in order to answer it successfully'', which makes it not challenging enough in terms of ``[c]ompositionality, and the corresponding ability to answer questions with arbitrarily complex structure''. To improve on that, they introduced the  \textsc{shapes} dataset which pairs synthetic images with synthetic, programmatically generated sentences that contain spatial relations. Two more datasets explored this direction \cite{Johnson2016,Suhr2017},
until another dataset \cite{Suhr2018} progressed beyond the use of synthetic images, in effect claiming that natural images are more valid for the task described as ``visual question answering''.

\subsection{Better Tasks and Games}
\label{sec:tasks}

We see a movement towards involvement of \emph{capabilities} already in the previous two sections. Where a model can be judged to be better than another one (trained on the same data) simply by the linear order imposed by the evaluation metric, a dataset must be argued to be more representative of a task by taking recourse to the capabilities of interest (e.g., in the example discussed just above, that of handling \emph{compositionality}).

As shown in Figure~\ref{fig:tag}, tasks are only grounded (to their left in the diagram) by capabilities (and games by tasks), and hence an argumentation for a task $T'$ in relation to a previous task $T$ should ideally make mention of how $C_{T'}$ and $C_T$ relate. (Similarly for games.)

\ \\
Figure~\ref{fig:prog} again summarises this discussion in a diagramm.

\subsection{From Models to Capabilities}
\label{sec:modcap}

An interesting additional avenue of research has been explored in recent years \cite[see for example the paper by][and references therein]{Hewitt2019}, where models trained on datasets (representing tasks or games) are explored for how they decomposed their (representational) task. For example, the question might be, as in the cited paper, whether a particular model trained on a particular task creates ``internally'' something akin to syntax trees. Again, why this might be interesting is typically not spelled out in these papers, but one can assume that the intended underlying argument goes something like this: ``If the theoretical construct is to be found, this shows that postulating it is empirically validated (and it can be learned simply through exposure to data); if it is not there, then the task can be done without it, to the extent that the model can do it, and it need not be postulated''.

\section{Conclusions: Making it Explicit}
\label{sec:conc}

The discussion above has mixed  normative (how it could, or should be) and descriptive (how it is) aspects. I will close by commenting more directly on what I think could be improved.

As discussed above, the most frequent way of contributing to the project is by improving models, for which established methods of evaluation exist. Next frequent is providing new or improved datasets. Here already we find a much larger variety of how the contribution is motivated and framed, and often it is taken as self-evident that a new dataset will drive progress. Here, more explicitness about the assumed advantages, and how they connect to the goal of modelling language competence, would be helpful.\footnote{%
  The recent initiative of formulating ``Data Sheets'' \cite{gebru:datasheets}, although developed for different purposes, would be a good step also in this direction.
}

This holds even more when introducing new tasks or games. As the discussion above hopefully has made plausible, motivating those in themselves, and relating them to existing ones, requires making claims about language capabilities. To make these in a convincing way, it seems advantageous to strive for a renewed, stronger connection to the sciences that study them: linguistics and cognitive psychology. Those fields provide the theoretical terms and constructs that would allow us to make the, as discussed above, typically  implicit arguments more explicit (and hence contestable).

\subsubsection*{Acknowledgements}
\label{acks}

{\small 
I thank Raquel Fern\'andez, Manfred Stede, and Sina Zarrie{\ss} for interesting discussions about topics related to this paper, while reserving the exclusive right to be blamed for any misunderstandings it might betray. \par
}

\bibliographystyle{acl_natbib}
\bibliography{/Users/das/work/projects/MyDocuments/BibTeX/all-lit.bib}


\appendix

\section{Formalising the Notions}
\label{sec:form}

\subsection{Tasks}
\label{sec:ftasks}

\begin{definition}
  A \emph{Language Task} is a tuple $(S, A, \mathcal{L}, D_T)$, where:
  \begin{itemize}[itemsep=.2\baselineskip,topsep=.5\baselineskip]
  \item $S$ is a (possibly infinite) set of \emph{states},
  \item $A$ is a (possibly infinite) set of \emph{actions},
  \item with either the states in $S$ or the actions in $A$ (or both) having as part \emph{natural language expressions}, and
  \item $\mathcal{L}: S \to A$ is a function that maps a state $s \in S$ to an action $a \in A$, where
  \item the mapping $\mathcal{L}$ conforms to \emph{task description} $D_T$.
  \end{itemize}
\end{definition}


\subsection{Worlds}
\label{sec:fworlds}

\begin{definition}
  A \emph{Micro-World} or \emph{Environment} is a tuple $(S, A, \mathcal{E}, R, D_W)$, where function $\mathcal{E}: S \times A \to S \times R$ maps an action $a$, taken in state $s$, to a state $s'$ and a reward $r$, and the mapping conforms to the \emph{world description} $D_W$.
\end{definition}

\subsection{Games}
\label{sec:fgames}

We approach the definition of a \emph{language game} via the more general notion of \emph{interaction game}:\footnote{%
  \emph{Pace} \citet{Witt:PU}, we get to define what a game is. Or we could go with \citet{suits:grasshopper}, who is happy to define games as rule-guided activities of voluntary attempt to overcome unnecessary obstacles. His concepts of \emph{prelusory goal}, which can be stated independently of the game (e.g., in football (soccer), ``make the ball be in the opponent team's goal''); \emph{constitutive rules}, which make reaching that goal more difficult than necessary (e.g., by disallowing to just grab the ball and carry it to the goal); and \emph{lusory attitude}, which is to accept the complications posed by the constitutive rules, can inform the design of games that work via crowdsourcing.
}

\begin{definition}
  An \emph{Interaction Game} is a tuple $(P, A, o, T, E, D_G)$, where:
\begin{itemize}[itemsep=.2\baselineskip,topsep=.5\baselineskip]

\item $P = \{p_1, \dots, p_n, N\}$ is the set of $n$ regular players $p$, together with one additional player $N$ (for \emph{Nature}).\\
  \emph{$N$ has a special status in that it does not have a strategic interest in the outcome of the game.}

\item $A = \{A_1, \dots, A_n, A_N\}$ is the set of action spaces, with one space per player.\\
  \emph{  Action types can be complex: e.g., $(nav, \mathrm{s})$, for ``navigation action, \emph{south}'', or $(utt, \mbox{``I don't know''})$ for ``utterance of \emph{I don't know}''. If defined in the right way (for example by a recursive grammar), the set of action types can be infinite.
    Players chose actions from their space of available actions; the resulting action tokens $a_j$ are associated with their originator through a function $\mathsf{a}: A_i \to P$, and with a position in the sequence of actions that have been performed since the beginning of the interaction through a function $\mathsf{t}: A_i \to \mathbb{N}$.
}

\item $o: P_i \times A_i \to \mathcal{P}(P)$ (for $P_i \in P, A_i \in A$) is the \emph{observability function} that specifies which types of actions by which player can be observed by which subset of the players.\\
  \emph{In normal cases, one would assume that players can observe their own actions, and that Nature observes all actions; but this allows for the specification of deviant cases.}

\item $T: \emptyset \cup (P \times A) \to \mathcal{P}(P)$ is the \emph{turn taking rule} that specifies who can act next, depending on who did what last. It also specifies who can start the game.\\
  \emph{  In a \emph{free initiative} setting, any player can act at any time; in a strict turn based setting, the current player would always be excluded from the set of next players.}
  
\item $E: S \to V$ is the \emph{evaluation rule} that maps a sequence of action tokens $\langle a_1, \dots, a_m\rangle$
  into an \emph{evaluation}, where the set of possible evaluations $V$ includes at least one positive one (e.g., \emph{success}) and one negative one (e.g., \emph{failure}).\\
  \emph{  The evaluation is made known to the players when a positive or negative outcome has been reached.
If it is not, or if it is does not contain outcomes denoted as positive or negative, we call the resulting structure an interaction \emph{setting}, rather than an interaction game.
}
\item $D_G$ finally is the \emph{game description} which specifies which, if any, otherwise existing activity the game is meant to approximate.
\end{itemize}
\end{definition}

The well-known \emph{Gridworld} game (see e.g.\ \citet{suttonbarto:rl}) for example can be represented in these terms as being an interaction game with one regular player (the agent) interacting with Nature, $P = \{p_1, N\}$. The agent can only perform navigation actions: $A_1 = \{(nav, \mathrm{n}), \dots\}$, Nature informs on the resulting available navigation options, $A_N = \{(\mathrm{\emph{inform}}, (n,w)), \dots\}$, with the information that Nature relays coming from a microworld that simulates the grid and the movement on it.

\emph{Gridworld} does not involve language, and hence is not an example of a \emph{language} interaction game. As an example of a language interaction setting that is not a game, we can define \emph{free chat interaction} in our terms as involving two players and an inert Nature that does not intervene: $P = \{p_1, p_2, N\}$, $A_1 = A_2 = \{(utt, \alpha)\}, A_N = \emptyset$, $T$ is a constant function into $P$ (free initiative), all actions are observed by all, $E$ is a constant function into $\{$\emph{undecided}$\}$.

\end{document}